\documentclass[10pt]{article}

\usepackage[margin=0.75in]{geometry}
\usepackage{amsmath,amssymb}
\usepackage{booktabs}
\usepackage{microtype}
\usepackage[hidelinks]{hyperref}
\usepackage{enumitem}
\usepackage{tikz}
\usetikzlibrary{arrows.meta,positioning,fit,calc,shapes.geometric}
\usepackage{array}
\usepackage{tabularx}
\usepackage{caption}
\usepackage{graphicx}
\usepackage{float}
\usepackage{xcolor}
\usepackage{setspace}
\usepackage{graphicx}
\usepackage{subcaption}
\title{\textbf{Traffic Sign Recognition for Autonomous Driving Using Branched YOLOv2 and Geometric Features}}

\author{
Arefeh Rezaei\\
Department of Computer Engineering,
K. N. Toosi University of Technology\\
\href{mailto:a_rezai@alumni.kntu.ac.ir}{a\_rezai@alumni.kntu.ac.ir}\\[6pt]
\textit{Supervisors:}\\
Alireza Fatehi \quad and \quad Behrooz Nasihatkon\\
\href{mailto:fatehi@kntu.ac.ir}{fatehi@kntu.ac.ir} \quad \href{mailto:nasihatkon@kntu.ac.ir}{nasihatkon@kntu.ac.ir}
}
\date{}

\begin{document}
\maketitle

\begin{center}
\small
\textit{Expanded English-language research version of the author's M.Sc. thesis submitted at K. N. Toosi University of Technology in 2019.}
\end{center}

\begin{abstract}
Traffic sign recognition (TSR) is an important perception task for autonomous driving and advanced driver-assistance systems because a useful system must determine both where a sign is located and which semantic class it belongs to, while operating under practical time constraints. This work presents a traffic sign recognition system built around YOLOv2 for simultaneous detection and classification. Two complementary modifications are studied. First, YOLOv2 is converted into a branched architecture with intermediate prediction layers. The branches make it possible to terminate computation for easy cases before the final network layers, thereby reducing inference time. Two branching rules are considered: termination based on the whole image and a cell-wise strategy that allows different feature-map cells to continue to different depths. Second, geometric information is introduced to reduce classification errors between visually similar signs. An unsupervised Bayesian image-segmentation procedure produces a binary representation from which class-specific geometric templates are constructed. The segmented region inside a YOLOv2 bounding box is compared with the template of the predicted class using a difference matrix. This geometric information is used either directly during inference or as an additional signal during parallel training of the detector. Because the experiments require samples containing both localization annotations and fine-grained sign labels, a dedicated dataset is constructed from GTSDB and GTSRB using seamless cloning and controlled image transformations. The reported experiments cover ten traffic-sign classes, with 3,000 training samples and 300 test samples. For the selected three-output branched architecture, the thesis reports a runtime of 0.647 s and mAP of 0.680, compared with 0.6607 s and 0.680 for the baseline YOLOv2. Using geometric features during inference raises the reported mAP to 0.713, while the geometric-feature training variant reaches 0.697 mAP at a reported runtime of 0.6608 s.
\end{abstract}


\section{Introduction}
Traffic sign recognition is a central component of the visual perception stack considered in autonomous-driving and advanced driver-assistance systems. A recognition module is expected to identify traffic signs in real-world scenes and provide a class label that can be used by downstream decision-making systems. The thesis motivating this paper treats the problem as a simultaneous detection-and-classification task, rather than separating localization and classification into independent pipelines.

Two practical requirements shape the design. The first is inference speed. An autonomous vehicle processes a continuous stream of frames, so unnecessary computation can delay the availability of a recognition result. The second is classification accuracy. Traffic signs can occupy a very small fraction of a road image, can be blurred or partially obscured, and can have similar colors and shapes across different classes. A detector may therefore localize a sign correctly while still assigning it to the wrong class.

The work uses YOLOv2 as a baseline because it formulates object detection as a unified regression problem and performs localization and classification in a single convolutional network. This structure avoids the sequential region-proposal and classification pipeline used by earlier detector families and provides a fast starting point for a real-time TSR system \cite{redmon2016,redmon2017}. The thesis then modifies the baseline in two directions: it adds intermediate prediction exits to reduce computation, and it supplements learned pixel-based features with explicit geometric information.

The two modifications are intentionally complementary. Branching changes \emph{how much} of the network a sample must traverse. Geometric processing changes \emph{what information} is available when a class decision is made. This gives the proposed system a design that addresses computational efficiency and recognition reliability separately while retaining a common YOLOv2 core.

A further challenge is data organization. Detection datasets generally provide object locations, whereas classification datasets provide fine-grained labels without the same scene context. The thesis therefore constructs a new training/testing set in which each image contains both a traffic-sign location and a class label suitable for simultaneous detection and classification. GTSDB supplies real-world detection scenes, while GTSRB supplies class-specific sign instances. Seamless cloning is used to combine them, after which additional image transformations increase the range of appearances presented during training.

The main contributions represented by the thesis work are summarized as follows:
\begin{enumerate}[leftmargin=*, itemsep=2pt]
    \item a branched version of YOLOv2 with intermediate exits for early termination;
    \item a cell-wise branching rule intended to preserve multiple detections in a single image;
    \item a geometric-feature mechanism based on unsupervised Bayesian image segmentation and class-specific templates;
    \item two ways of using the geometric information, during inference or during parallel training; and
    \item a generated dataset supporting simultaneous traffic-sign detection and fine-grained classification.
\end{enumerate}

\section{Related Work and Design Motivation}
\subsection{Traffic Sign Recognition}
Traffic sign recognition has been studied with both hand-crafted features and deep convolutional networks. Earlier traffic-sign systems considered region proposals, color or shape cues, histogram-of-oriented-gradient features, support-vector machines, sliding windows, and cascaded classifiers. The thesis surveys multiple such approaches and uses them to motivate a shift toward unified deep detectors that can operate at higher speed.

The thesis also compares several object-detection families for traffic-sign detection, including region-based methods, SSD, and YOLO. The comparison is not limited to detection accuracy: it considers mAP, runtime, memory requirements, floating-point operations, model size, and the influence of sign size. In the surveyed experiments, YOLOv2 and SSD were among the fastest methods, while SSD was reported as less suitable for small traffic-sign objects in the specific comparison used by the thesis. The work consequently selects YOLOv2 as the detector to be improved.

\subsection{Unified Detection with YOLO}
YOLO's key design choice is to treat detection as a single regression task. Instead of first proposing regions and then applying a separate classifier, the network receives the entire image and directly predicts bounding boxes and class probabilities. This gives the model global image context and avoids repeated feature extraction for multiple proposals \cite{redmon2016}.

YOLOv2 extends the initial design with batch normalization, higher-resolution classification pretraining, anchor boxes, and further architectural changes that improve recall and localization while retaining a fast inference path \cite{redmon2017}. These properties make YOLOv2 an appropriate baseline for a system in which runtime matters as much as raw recognition accuracy.

\subsection{Branching as a Computational Strategy}
The thesis draws on earlier work on branched convolutional networks for traffic-sign recognition. The motivating principle is that examples of different complexity do not necessarily need the same depth of processing. If a sample is easy, a shallow representation can be sufficient; if it is difficult, deeper features can be used. The present work transfers this concept from classification toward the simultaneous detection-and-classification setting by placing intermediate outputs inside YOLOv2.

A major design issue is the unit of early termination. Terminating a whole image after a single confident detection can save computation but may suppress other signs in the same frame. The cell-wise design introduced in this work therefore treats feature-map cells independently, so that one easy sign does not automatically terminate the processing of other cells.

\subsection{Shape-Based Information}
The thesis further adopts the idea that object shape can provide information not fully captured by a purely pixel-based representation. An object-based convolutional approach for high-resolution imagery is used as a conceptual reference for combining object-level and pixel-level information \cite{zhao2017}. For traffic signs, this motivation is especially relevant because multiple classes can share similar color distributions while differing in geometric structure.

The geometric path in the present system is implemented using fast unsupervised Bayesian image segmentation with adaptive spatial regularization \cite{pereyra2017}. The segmentation output is converted into class-specific templates, and these templates provide a shape-consistency signal that can be combined with YOLOv2's class prediction.

\section{Baseline YOLOv2 Formulation}
\subsection{Unified Detection Representation}
In the YOLO formulation described by the thesis, the input image is divided into an $S\times S$ grid. Each cell predicts a fixed number $B$ of bounding boxes. Each predicted bounding box contains its geometric parameters and an objectness/confidence value, while each cell also contributes class probabilities. The resulting output jointly represents localization and classification.

For a bounding box, the normalized parameters are the center coordinates and dimensions, together with a confidence quantity. The thesis defines the confidence using the probability that an object exists in the box and the intersection-over-union (IoU) between the predicted box and the reference box. IoU is computed as
\begin{equation}
\mathrm{IoU}=\frac{\mathrm{area}(B_{pred}\cap B_{gt})}{\mathrm{area}(B_{pred}\cup B_{gt})}.
\end{equation}
At inference time, a ground-truth box is not available, so the network's predicted confidence is used directly for decision making.

For each cell, the class score of a predicted box can be viewed conceptually as the product of the objectness estimate and the conditional class probability. Redundant boxes are then filtered with non-maximum suppression (NMS), so that highly overlapping predictions of the same class do not remain simultaneously active.

\subsection{Training Objective}
The thesis describes the YOLO loss as a sum of localization, confidence, and classification components. The localization term penalizes differences between predicted and reference box coordinates and dimensions. The confidence term separates boxes responsible for objects from background boxes. The classification component penalizes incorrect class predictions for cells responsible for an object.

The objective can therefore be expressed generically as
\begin{equation}
\mathcal{L}=\mathcal{L}_{loc}+\mathcal{L}_{obj}+\mathcal{L}_{noobj}+\mathcal{L}_{cls},
\end{equation}
where the localization term is weighted to emphasize accurate box geometry and the background-confidence term is downweighted to avoid the many empty cells dominating the loss. In the YOLO formulation summarized in the thesis, $\lambda_{coord}=5$ and $\lambda_{noobj}=0.5$ are used for this balancing.

\subsection{Why YOLOv2 Is Retained}
The proposed research does not replace YOLOv2 with a separate detector. Instead, it uses YOLOv2 as the central representation and modifies the execution path and decision signals around it. This choice is important because the final application is assumed to be a real-time perception task. A redesign that significantly increases the cost of the baseline would undermine the main deployment objective.

\section{Proposed Branched YOLOv2}
\subsection{Architecture}
\begin{figure}[H]
\centering
\resizebox{0.96\linewidth}{!}{%
\begin{tikzpicture}[
    node distance=6mm and 6mm,
    box/.style={draw, rounded corners, align=center, minimum width=29mm, minimum height=9mm, font=\small},
    smallbox/.style={draw, rounded corners, align=center, minimum width=25mm, minimum height=7mm, font=\scriptsize},
    arrow/.style={-{Latex[length=2.2mm]}, thick},
    group/.style={draw, rounded corners, inner sep=4mm, dashed}
]
\node[box] (input) {Input road image};
\node[box, right=of input] (yolo) {YOLOv2\\feature extraction};
\node[smallbox, above right=of yolo, xshift=8mm] (e1) {Intermediate\\output 1};
\node[smallbox, right=of yolo, xshift=8mm] (e2) {Intermediate\\output 2};
\node[smallbox, below right=of yolo, xshift=8mm] (e3) {Intermediate\\output 3};
\node[smallbox, right=of e2, xshift=8mm] (final) {Final YOLOv2\\output};
\node[box, below=15mm of e2] (seg) {Unsupervised Bayesian\\image segmentation};
\node[smallbox, right=of seg, xshift=7mm] (crop) {Predicted-box\\segmented region};
\node[smallbox, right=of crop, xshift=7mm] (templ) {Class-specific\\geometric template};
\node[smallbox, right=of templ, xshift=7mm] (fusion) {Geometric verification\\or fusion};
\draw[arrow] (input) -- (yolo);
\draw[arrow] (yolo) -- (e1);
\draw[arrow] (yolo) -- (e2);
\draw[arrow] (yolo) -- (e3);
\draw[arrow] (e1) -- ++(0.8,0) |- (final);
\draw[arrow] (e2) -- (final);
\draw[arrow] (e3) -- ++(0.8,0) |- (final);
\draw[arrow] (yolo) |- (seg);
\draw[arrow] (seg) -- (crop);
\draw[arrow] (crop) -- (templ);
\draw[arrow] (templ) -- (fusion);
\draw[arrow] (final) -- (fusion);
\node[group, fit=(e1)(e2)(e3), label={[font=\scriptsize]above:branched / early-exit path}] {};
\node[group, fit=(seg)(crop)(templ)(fusion), label={[font=\scriptsize]below:shape-based accuracy path}] {};
\end{tikzpicture}%
}
\caption{Schematic architecture of the proposed system. Intermediate outputs provide early decisions, while the parallel segmentation path supplies explicit geometric information. The diagram follows the structure described in the thesis and is not a direct copy of the original thesis figure.}
\label{fig:architecture}
\end{figure}
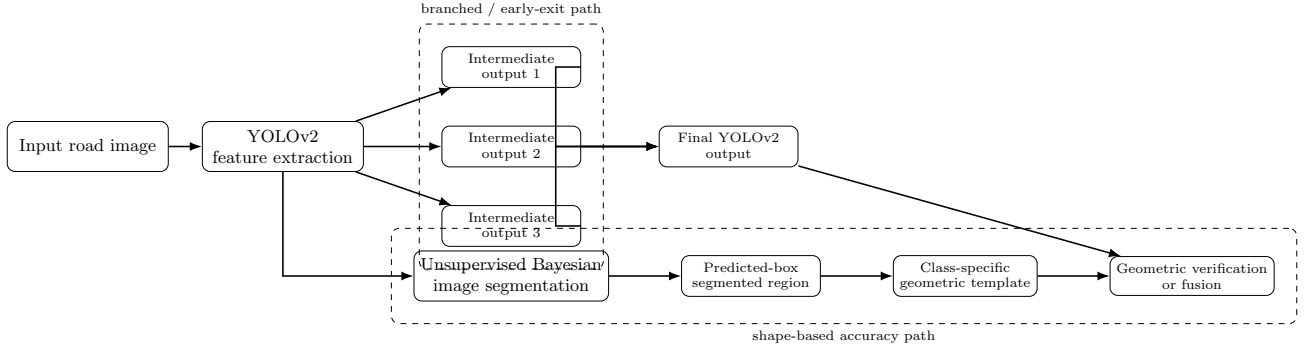

Figure~\ref{fig:architecture} summarizes the architecture reconstructed from the thesis description. The YOLOv2 backbone remains the main feature extractor. Three intermediate prediction outputs are introduced at selected positions. At each output, part of the current representation can be converted into a detection/classification decision. Samples or cells that meet the early-exit condition leave the deeper computation, while uncertain cases continue toward later outputs.

\subsection{Choosing the Number and Location of Exits}
The thesis evaluates multiple configurations rather than assuming that more exits are always better. Three-, four-, and five-output variants are trained with different locations. The reason for this search is that an intermediate output should remove a meaningful amount of computation. If two exits are placed too close to each other, the second can inherit little additional benefit and can instead add training and inference overhead.

Another consideration is the interaction with max-pooling. The thesis initially examines placements before the fourth and fifth max-pooling layers. Those placements do not improve speed in the experiments because they increase computation without providing sufficiently early useful exits. The final selected configuration places the three intermediate outputs after max-pooling operations, outside the DarkNet backbone, in the arrangement identified experimentally as the most favorable.

\begin{table}[H]
\centering
\caption{Reported evaluation of 2YOLOv-B with different numbers and placements of intermediate output layers.}
\label{tab:branch_layers}
\small
\begin{tabular}{lcc}
\toprule
Configuration & Runtime (s) & mAP\\
\midrule
Three layers, placement 1 & 0.6559 & 0.680\\
Three layers, placement 2 & 0.6470 & 0.680\\
Four layers & 0.6708 & 0.680\\
Five layers & 0.6760 & 0.680\\
\bottomrule
\end{tabular}
\end{table}

The selected three-output configuration is therefore the second placement in Table~\ref{tab:branch_layers}. The thesis reports the same mAP across the tested alternatives, while the runtime varies with the placement and number of intermediate outputs.

\subsection{Whole-Image Branching}
The first branching implementation makes an early decision using the complete image. If an intermediate output produces a box whose sign probability exceeds 0.5, the sign is reported and the image is removed from further processing.

This mechanism is straightforward and can terminate easy examples early. However, it creates an image-level coupling between detections. Suppose an image contains one sign that is easy and another that is difficult. If the easy sign triggers an early exit, the entire image may leave the network before the difficult sign reaches a sufficiently deep representation. The thesis identifies this case as a failure mode of whole-image branching.

\subsection{Cell-Wise Branching}
The final implementation moves the decision to the level of feature-map cells. The method evaluates the probability of object presence and the class probability associated with each cell. The thesis uses two thresholds. With a high threshold of 0.5, a sufficiently confident detection can be accepted and the corresponding cell can leave the computation. With a low threshold of 0.1, a cell whose probability is too small can be discarded. Intermediate cases continue into deeper layers.

The decision rule can be summarized as
\begin{equation}
\mathrm{state}(p)=
\begin{cases}
\mathrm{exit}, & p\geq 0.5,\\
\mathrm{discard}, & p<0.1,\\
\mathrm{continue}, & 0.1\leq p<0.5.
\end{cases}
\end{equation}
This rule is applied at the cell level, so one cell can exit while neighboring cells continue. The thesis emphasizes two sources of saved computation: low-probability cells are removed from later processing, and high-confidence cells terminate early.

The reported comparison between the two branching implementations is shown in Table~\ref{tab:branch_compare}. The whole-image version has a runtime of 0.647 s, while the cell-wise version has a runtime of 0.6527 s. Both retain the reported mAP of 0.680. Although the whole-image rule is slightly faster in this measurement, the cell-wise rule addresses the multi-sign failure mode described above and is therefore the conceptually relevant strategy for scenes containing several signs.

\begin{table}[H]
\centering
\caption{Reported comparison of the YOLOv2 baseline and the two 2YOLOv-B branching strategies.}
\label{tab:branch_compare}
\small
\begin{tabular}{lcc}
\toprule
Model & Runtime (s) & mAP\\
\midrule
YOLOv2 & 0.6607 & 0.680\\
2YOLOv-B, whole-image branching & 0.6470 & 0.680\\
2YOLOv-B, cell-wise branching & 0.6527 & 0.680\\
\bottomrule
\end{tabular}
\end{table}

\subsection{Feature Fusion at Intermediate Outputs}
The intermediate predictions must operate with less information than the final output. To strengthen these exits, the thesis combines the features at intermediate outputs with features from earlier layers inside the DarkNet structure. The motivation is to provide the shallow prediction heads with both their current representation and information retained from preceding stages.

The resulting experiment compares intermediate outputs trained without feature fusion against the same outputs trained with fusion. The reported mAP increases from 0.680 to 0.689 while runtime changes from 0.6470 s to 0.6491 s. The small runtime difference is consistent with the design goal: improve the quality of early predictions without eliminating the computational advantage of branching.

\begin{table}[H]
\centering
\caption{Reported effect of feature fusion for 2YOLOv-B intermediate outputs.}
\label{tab:fusion}
\small
\begin{tabular}{lcc}
\toprule
2YOLOv-B configuration & Runtime (s) & mAP\\
\midrule
Without intermediate feature fusion & 0.6470 & 0.680\\
With feature fusion & 0.6491 & 0.689\\
\bottomrule
\end{tabular}
\end{table}

\section{Geometric-Feature Enhancement}
\subsection{Motivation}
Detection and localization alone do not guarantee a correct traffic-sign class. The thesis observes that signs with related visual appearances can be confused by a purely learned pixel representation. The proposed solution is to provide an additional feature family describing geometric structure.

The design deliberately avoids a large secondary recognition network. Instead, the geometric path is built from a segmentation result and a small set of class-specific templates. This gives a relatively compact auxiliary representation that can be compared with a detected sign region.

\subsection{Unsupervised Bayesian Image Segmentation}
The segmentation component follows the fast unsupervised Bayesian image-segmentation framework with adaptive spatial regularization described by Pereyra and McLaughlin \cite{pereyra2017}. In the formulation used by the thesis, image pixels are associated with latent classes, and the class assignments are coupled spatially through an MRF/Potts prior. The regularization parameter is not fixed manually as an external constant; instead, the inference procedure integrates it out and obtains an adaptive formulation.

The derivation in the thesis introduces an auxiliary variable and uses a small-variance asymptotic analysis. The resulting optimization problem contains a data-fidelity term and a spatial regularization term. A non-convex $\ell_0$ gradient term is relaxed with total variation so that the problem can be solved through a sequence of convex subproblems. The implementation iterates between a TV-based denoising/update step and a clustering step based on least-squares clustering (k-means).

For the application in this paper, the important output is not the latent statistical interpretation itself but the resulting binary segmented image. This image provides a simplified representation of sign geometry that is less dependent on the original RGB texture.

\subsection{Class-Specific Templates}
For each traffic-sign class, a binary geometric template is created. When a traffic sign is detected by YOLOv2, its predicted bounding box is mapped to the segmented image. The region inside the predicted box is extracted and resized to the template dimensions.

Let $T_c$ denote the template of class $c$ and $S$ the segmented region extracted from a YOLOv2 bounding box. A difference matrix can be defined elementwise as
\begin{equation}
D_c(i)=|S(i)-T_c(i)|,
\end{equation}
and the scalar comparison value used by the thesis is the mean of this matrix,
\begin{equation}
d_c=\frac{1}{N}\sum_{i=1}^{N}D_c(i).
\end{equation}
The resulting score measures how well the geometric structure inside the predicted box matches the expected shape of the predicted class.

\subsection{Black-White Inversion}
Binary segmentation can produce the same shape with black and white values swapped. Direct comparison to a fixed template could therefore give a large difference even when the geometric structure is correct.

The thesis addresses this by considering both the original template and the inverted binary representation. Experimental observations show that same-class comparisons occupy two near-extreme ranges: approximately $[0,0.27]$ without black-white inversion and approximately $[0.72,1]$ with inversion. Non-matching signs and non-sign image regions generally produce values in an intermediate range. The thesis therefore treats these two near-extreme intervals as consistent with the template and the middle region as inconsistent.

\begin{figure}[H]
\centering
\resizebox{0.94\linewidth}{!}{%
\begin{tikzpicture}[
    node distance=8mm and 8mm,
    box/.style={draw, rounded corners, align=center, minimum width=30mm, minimum height=10mm, font=\small},
    arrow/.style={-{Latex[length=2.3mm]}, thick}
]
\node[box] (rgb) {RGB image};
\node[box, right=of rgb] (seg) {Bayesian\\segmentation};
\node[box, right=of seg] (region) {Crop predicted\\bounding-box region};
\node[box, above right=of region, xshift=4mm] (temp) {Template\\for predicted class};
\node[box, below right=of region, xshift=4mm] (inv) {Inverted\\template};
\node[box, right=30mm of region] (diff) {Difference matrix\\and mean score};
\node[box, right=of diff] (dec) {Shape consistency\\decision};
\draw[arrow] (rgb) -- (seg);
\draw[arrow] (seg) -- (region);
\draw[arrow] (region) -- (diff);
\draw[arrow] (temp) -- (diff);
\draw[arrow] (inv) -- (diff);
\draw[arrow] (diff) -- (dec);
\draw[arrow] (region) |- (temp);
\draw[arrow] (region) |- (inv);
\end{tikzpicture}%
}
\caption{Geometric-feature pipeline reconstructed from the thesis: segmentation, region extraction, comparison with original and inverted templates, and shape-consistency decision.}
\label{fig:geometric}
\end{figure}
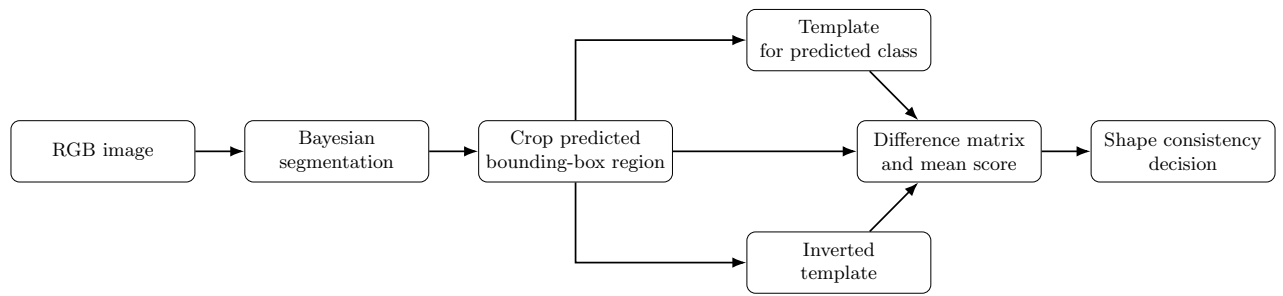

\subsection{Inference-Time Geometric Verification}
The simplest geometric variant leaves YOLOv2 training unchanged. During inference, the class predicted by YOLOv2 is checked against the geometric template. If the shape score is consistent with the predicted class, the result can be displayed or accepted; if it is inconsistent, the prediction can be rejected or corrected according to the decision logic of the implementation.

An attractive property of this mode is that the templates are reusable. The thesis explicitly notes that real-world input images are seen for the first time during operation and do not have annotation files, but the precomputed class templates can still be applied to those new images. Thus, geometric verification does not require ground-truth annotations at deployment time.

\subsection{Parallel Training with Geometric Features}
The second geometric variant incorporates the shape information during training. Each training image and its annotation are passed to YOLOv2 and the Bayesian segmentation component in parallel. The detected region is aligned with the template associated with the ground-truth class, and the resulting difference information is combined with the class probability from YOLOv2.

This makes the network weights sensitive to the geometric signal. The computational implication is a deliberate shift: training becomes more expensive because the auxiliary pipeline must run throughout optimization, but inference can remain close to the baseline because the network has already learned from the geometric information. The thesis labels this configuration 2YOLOv2-O.

\section{Dataset Construction}
\subsection{Need for a New Dataset}
The thesis identifies a structural mismatch between detection and classification data. Detection samples specify where an object is located, whereas classification samples provide the fine-grained identity of an object. A simultaneous detector/classifier requires both kinds of information for the same training example.

The solution is to combine the German Traffic Sign Detection Benchmark (GTSDB) and the German Traffic Sign Recognition Benchmark (GTSRB). GTSDB provides scene-level detection images and localization annotations, while GTSRB provides examples of traffic signs associated with individual classes \cite{houben2013,stallkamp2011}.

\subsection{Seamless Cloning Procedure}
The dataset-generation pipeline starts from a GTSDB scene. The original sign location is obtained from its annotation. A sign instance from GTSRB is then inserted into that location using seamless cloning. For each GTSDB detection image, multiple generated images can therefore be created, one for each desired sign class.

The thesis uses the object location already present in the detection annotation and changes the identity of the sign occupying that location. A new VOC-style XML annotation is generated for each image, preserving the bounding-box information and assigning the appropriate class name.

Seamless cloning is used because it blends image gradients across the pasted region and produces a more natural insertion than simple cut-and-paste. This makes the process suitable for creating a large number of training images programmatically.

\subsection{Data Augmentation}
The generated samples are further modified to introduce appearance variation. The thesis reports gamma correction for brighter and darker scenes, left/right rotations of signs, and affine transformations. The transformations are distributed approximately uniformly across the sign classes. The motivation is to avoid systematic under-representation of a particular variation within one class.

\begin{table}[H]
\centering
\caption{Reported dataset configuration.}
\label{tab:dataset}
\small
\begin{tabularx}{\linewidth}{>{\raggedright\arraybackslash}p{0.43\linewidth} X}
\toprule
Item & Reported value\\
\midrule
Detection source & GTSDB\\
Classification source & GTSRB\\
Generated samples & More than 10,000\\
Samples used in reported experiments & 3,300\\
Number of classes & 10\\
Training samples per class & 300\\
Test samples per class & 30\\
Total training samples & 3,000\\
Total test samples & 300\\
Original image size & $1360\times800$ pixels\\
Annotation format & VOC-style XML\\
\bottomrule
\end{tabularx}
\end{table}

The ten classes used in the experiments include four speed signs (30, 70, 80, and 100), two triangular signs (danger and pedestrian), and four directional signs (left turn, right turn, straight, and right-or-straight), as described in the thesis.

\section{Implementation and Experimental Setup}
\subsection{Software and Hardware}
The YOLOv2 implementation uses the Lightnet library on top of PyTorch. The thesis initially implemented YOLOv2 in TensorFlow but found the training and testing speed unsuitable for the target application, motivating a change of framework. The Bayesian segmentation algorithm was also implemented in PyTorch so that tensor and matrix information could be exchanged between the two processing paths.

The reported training environment consisted of a server with one NVIDIA GTX 1080 GPU and two 730-series GPUs, Ubuntu 18.04 (64-bit), approximately 31.3 GB of memory, 475.3 GB of disk storage, and an Intel Xeon E5-2620 v4 CPU at 2.10 GHz with 32 logical cores.

\subsection{YOLOv2 Training Parameters}
The thesis distinguishes parameters specified before execution from parameters learned during training. The reported fixed configuration includes the class labels, input image dimensions, batch settings, image/annotation paths, preprocessing parameters, and dataset-specific anchor boxes. The confidence threshold is 0.001 and the NMS threshold is 0.5. Training uses 160 epochs, an initial learning rate of $10^{-3}$, and momentum of 0.9. Model weights are stored after each 5,000 training batches.

\begin{table}[H]
\centering
\caption{Reported experimental settings for YOLOv2 and Bayesian segmentation.}
\label{tab:settings}
\small
\begin{tabular}{ll}
\toprule
Parameter & Reported setting\\
\midrule
Training epochs & 160\\
Initial learning rate & $10^{-3}$\\
Momentum & 0.9\\
YOLO confidence threshold & 0.001\\
NMS threshold & 0.5\\
Weight-save interval & 5,000 batches\\
Bayesian outer iterations & 15\\
Bayesian inner iterations & 5\\
TV stopping parameter & $10^{-4}$\\
Segmentation clusters & 2\\
\bottomrule
\end{tabular}
\end{table}

\subsection{Training Time}
The thesis reports approximately one week as a suitable training duration for YOLOv2; longer training could lead to overfitting in the reported setting. Branched models require additional training because all branches must be trained with the available data. The three-intermediate-output configuration requires approximately ten days, while four- and five-output structures take approximately two weeks.

This difference illustrates an important characteristic of the proposed strategy. The branching mechanism is intended to reduce inference cost, not training cost. The deeper and more complex the branching structure becomes, the more training effort is required.

\subsection{Segmentation and Template Thresholds}
The reported Bayesian segmentation configuration uses two pixel clusters. The template-matching threshold is derived empirically from the distribution of difference-matrix means on the experimental dataset. Same-class pairs occupy the near-zero and near-one intervals described above, whereas mismatched signs and non-sign regions fall largely within the intermediate interval.

This thresholding rule is therefore not claimed as a universal constant. It is a parameter chosen from the thesis experiments and should be treated as part of the reported experimental configuration.

\section{Experimental Results}
\subsection{Branching Results}
The intermediate-output experiments show that the number and placement of exits influence runtime even when mAP remains approximately constant in the reported tests. The three-output placement selected in the thesis achieves 0.6470 s runtime at 0.680 mAP.

The whole-image and cell-wise strategies are then compared. Their reported mAP values remain at 0.680, while runtime is 0.6470 s and 0.6527 s, respectively. The cell-wise rule is retained as the conceptually safer design for scenes containing multiple signs because it does not terminate every feature-map cell merely because one easy sign has already been recognized.

\subsection{Intermediate Feature Fusion}
Adding earlier features to the intermediate outputs raises the reported mAP from 0.680 to 0.689, while runtime increases modestly from 0.6470 s to 0.6491 s. This experiment suggests that early exits can benefit from feature fusion without eliminating the main speed motivation of the branched architecture.

\subsection{Geometric-Feature Results}
Table~\ref{tab:accuracy} summarizes the principal accuracy experiments. The baseline YOLOv2 reports 0.680 mAP at 0.6607 s. Applying geometric features only during inference produces 0.713 mAP at 0.6714 s. Training the network with the geometric signal (2YOLOv2-O) gives 0.697 mAP at 0.6608 s.

\begin{table}[H]
\centering
\caption{Reported performance of YOLOv2 and geometric-feature variants.}
\label{tab:accuracy}
\small
\begin{tabular}{lcc}
\toprule
Model / strategy & Runtime (s) & mAP\\
\midrule
YOLOv2 & 0.6607 & 0.680\\
Geometric features at inference & 0.6714 & 0.713\\
2YOLOv2-O, trained with geometric features & 0.6608 & 0.697\\
\bottomrule
\end{tabular}
\end{table}

The reported numbers show a clear trade-off between extra inference processing and model-integrated geometric information. The inference-time strategy adds an explicit shape-comparison step and therefore has the largest runtime among the three rows. The trained geometric variant instead moves the shape information into the learned model and remains close to the baseline runtime.

\subsection{Qualitative Behavior}
The thesis includes qualitative examples for all three main configurations. The examples show that the systems can recognize small signs relative to the image dimensions, including cases where the sign is easy to see and cases where it is harder to distinguish. One example shows a right-turn sign being detected at the second intermediate output of the branched network. Other examples illustrate classes that are correctly assigned by the geometric-feature variant after being assigned to incorrect classes by the baseline YOLOv2.

\textbf{Qualitative Results}: Figure X presents representative qualitative results obtained from the proposed traffic sign recognition system on real-world road scenes. The examples include traffic signs of different sizes and appearances under varying visual conditions, including relatively clear and challenging scenes. The results illustrate the ability of the proposed YOLOv2-based framework to localize traffic signs while simultaneously assigning their corresponding classes. In particular, the branched architecture enables some signs to be recognized at intermediate stages of the network, while the geometric-feature-based approach provides additional information for distinguishing visually similar signs and reducing classification errors. These examples complement the quantitative evaluation by demonstrating the behavior of the proposed methods on realistic traffic scenes.

\begin{figure*}[!t]
    \centering

    \begin{subfigure}[b]{0.48\textwidth}
        \centering
        \includegraphics[width=\linewidth]{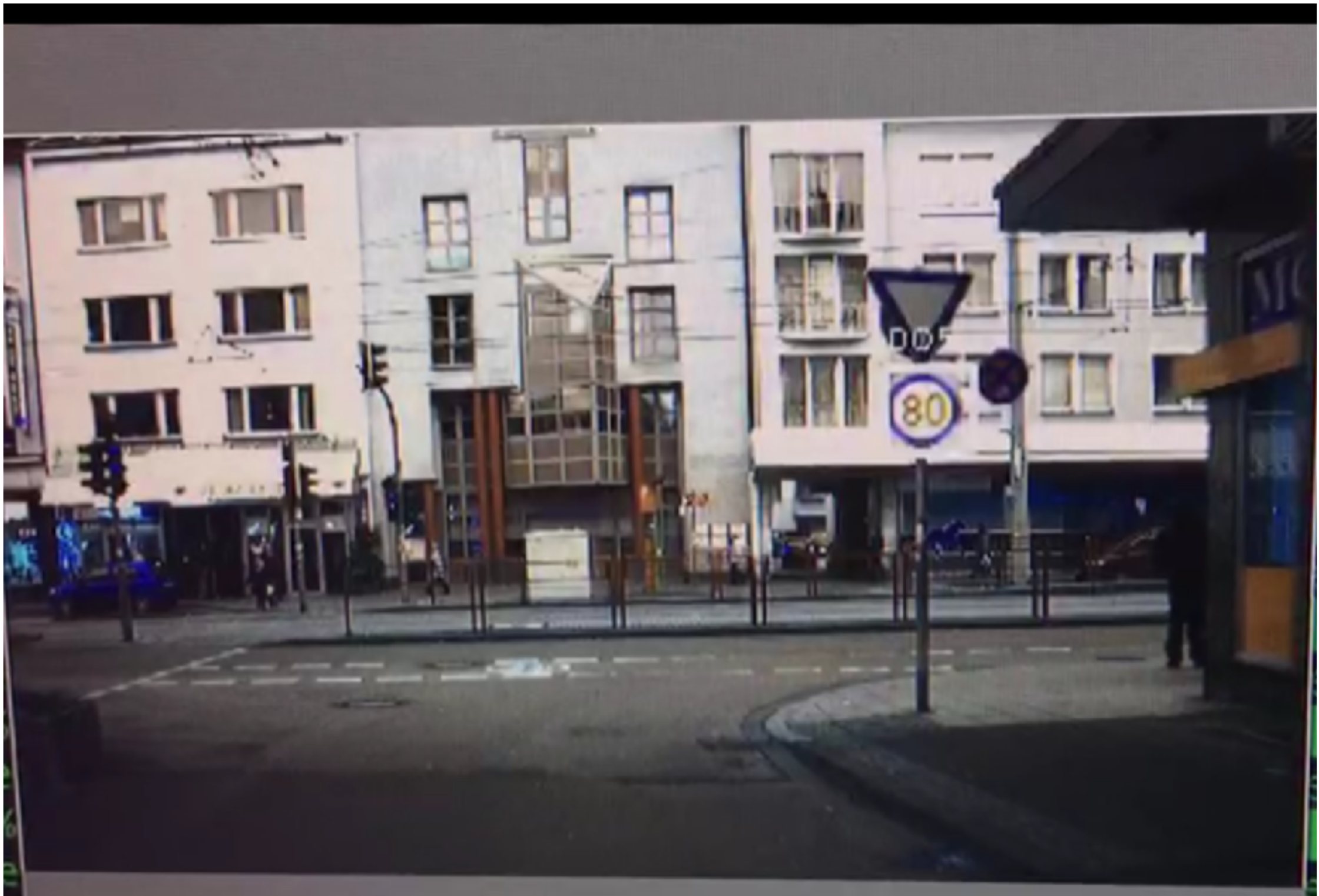}
        \caption{}
        \label{fig:result1}
    \end{subfigure}
    \hfill
    \begin{subfigure}[b]{0.48\textwidth}
        \centering
        \includegraphics[width=\linewidth]{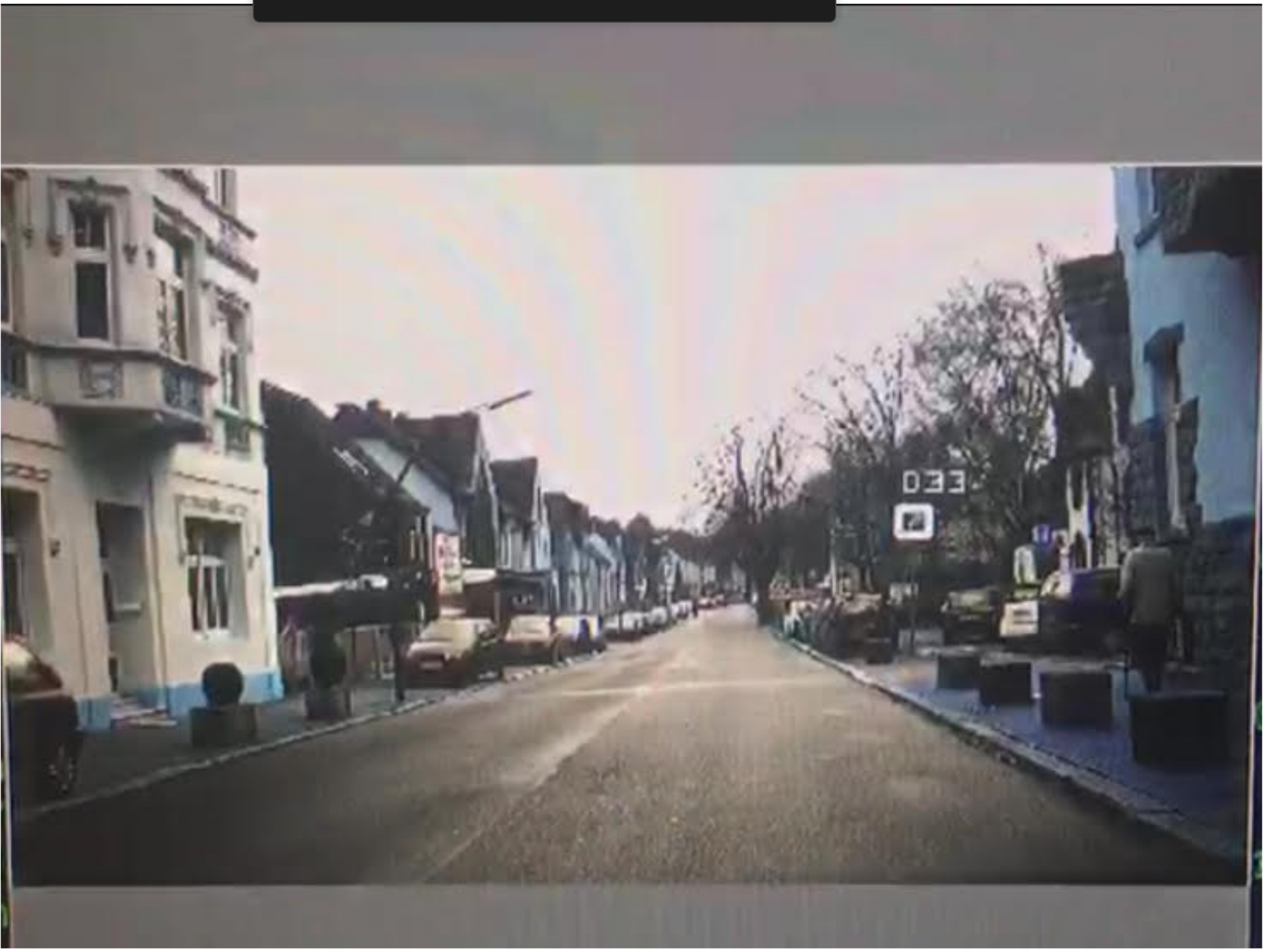}
        \caption{}
        \label{fig:result2}
    \end{subfigure}

    \vspace{0.15cm}

    \begin{subfigure}[b]{0.48\textwidth}
        \centering
        \includegraphics[width=\linewidth]{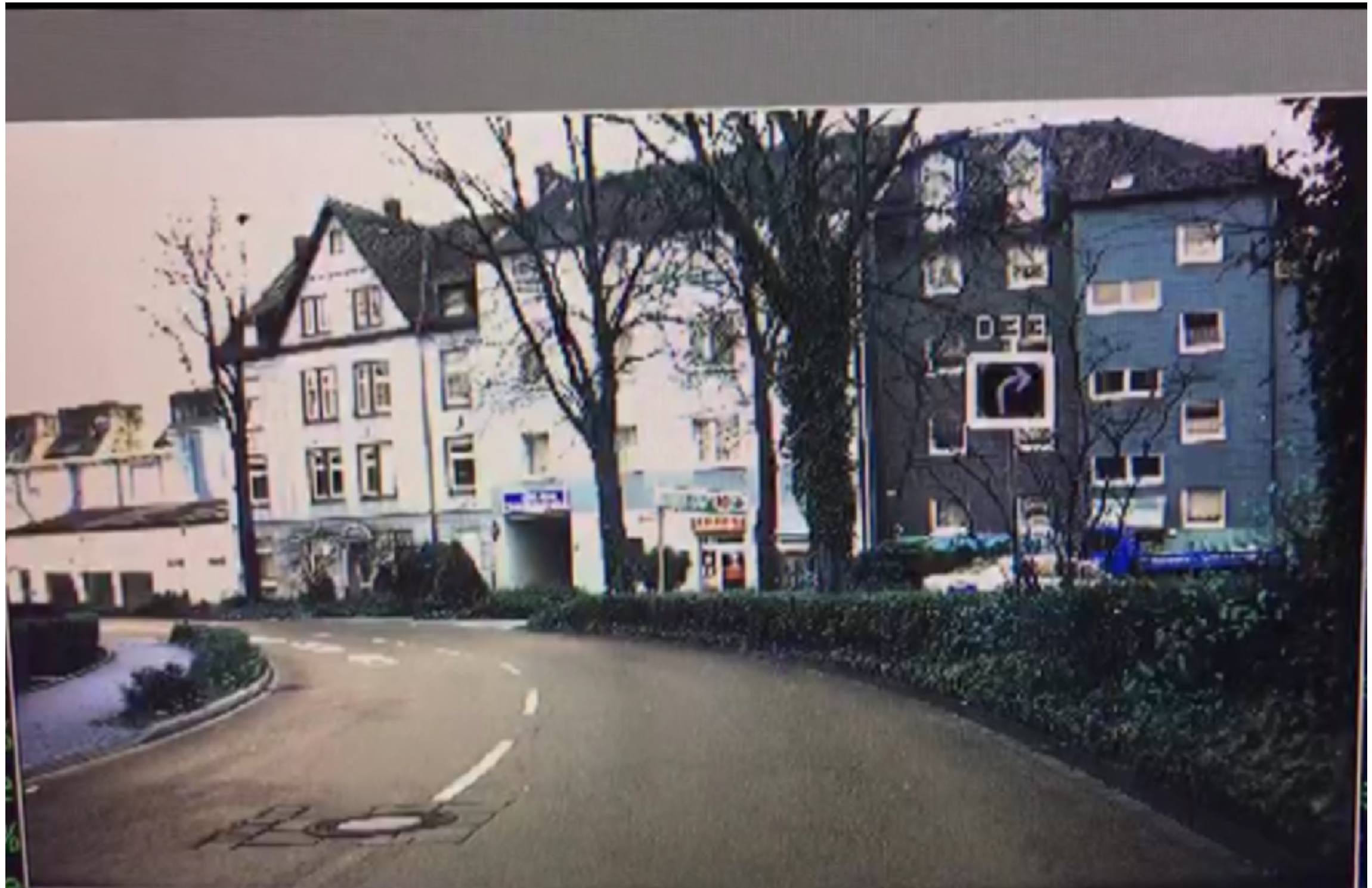}
        \caption{}
        \label{fig:result3}
    \end{subfigure}
    \hfill
    \begin{subfigure}[b]{0.48\textwidth}
        \centering
        \includegraphics[width=\linewidth]{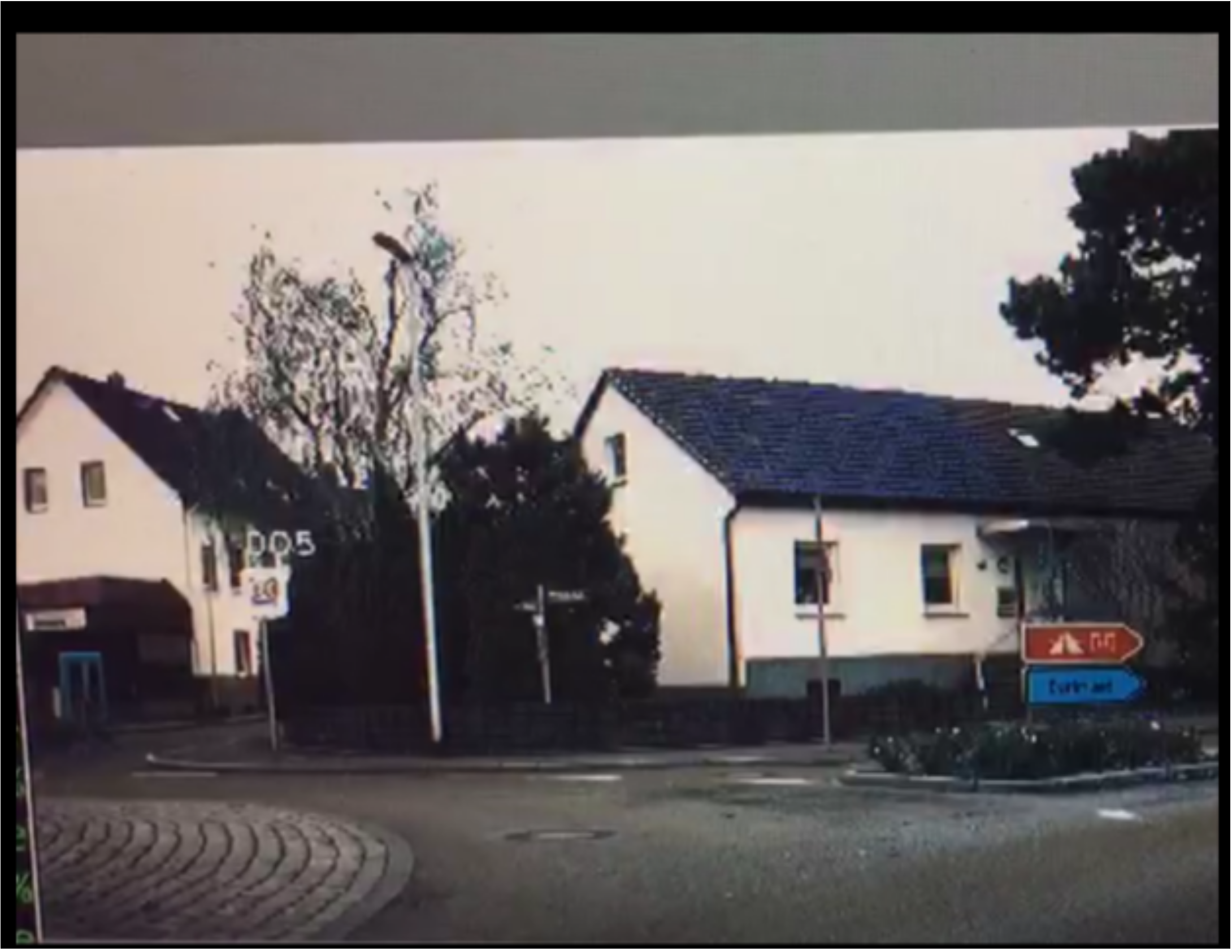}
        \caption{}
        \label{fig:result4}
    \end{subfigure}

    \vspace{0.15cm}

    \begin{subfigure}[b]{0.48\textwidth}
        \centering
        \includegraphics[width=\linewidth]{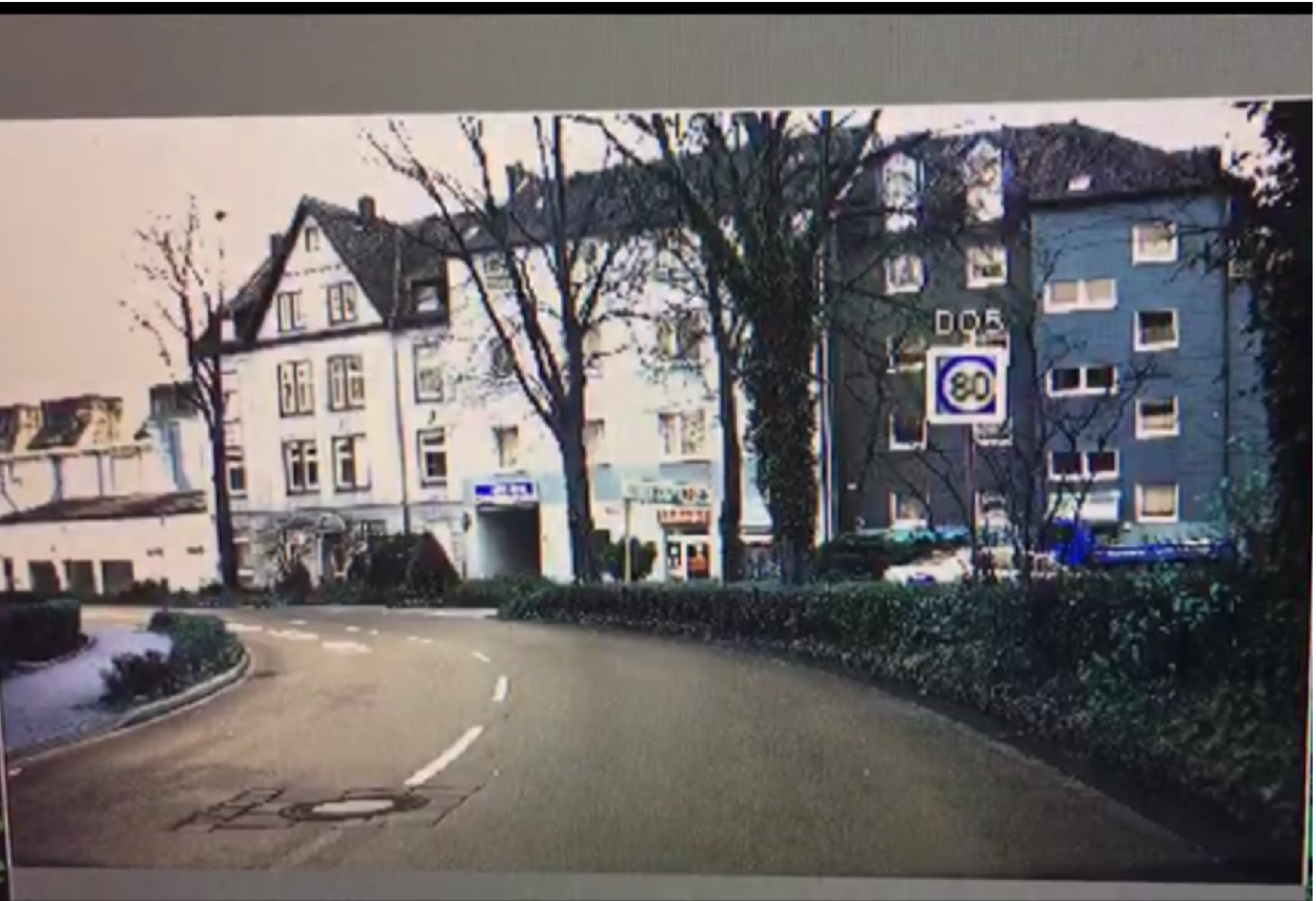}
        \caption{}
        \label{fig:result5}
    \end{subfigure}
    \hfill
    \begin{subfigure}[b]{0.48\textwidth}
        \centering
        \includegraphics[width=\linewidth]{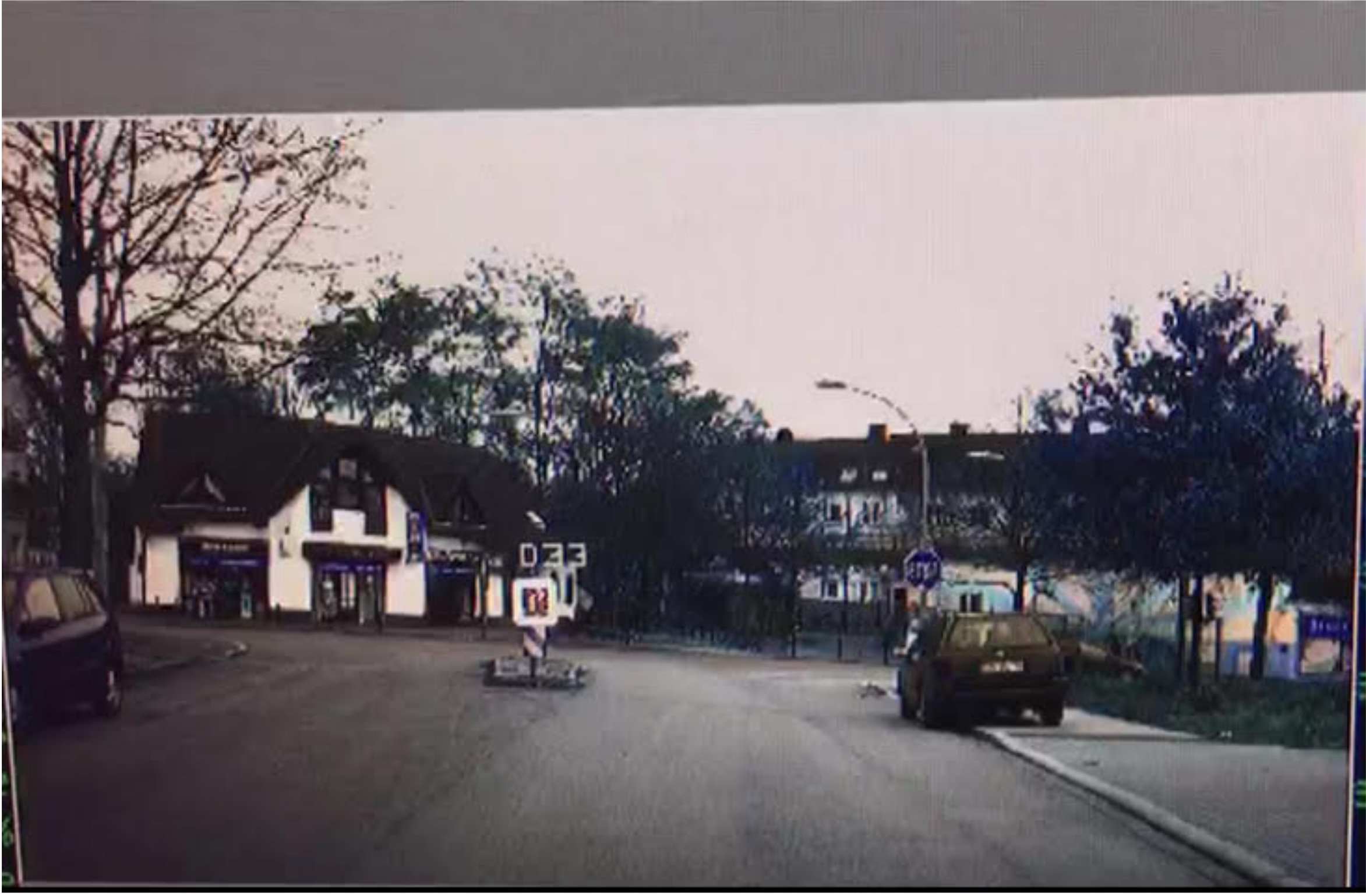}
        \caption{}
        \label{fig:result6}
    \end{subfigure}

    \caption{Representative qualitative results of the proposed traffic sign
    recognition system on real-world road scenes under different visual
    conditions. The examples illustrate traffic sign detection and
    classification using the proposed YOLOv2-based framework and its variants.}
    \label{fig:qualitative_results}
\end{figure*}

\subsection{Derived Comparison of Reported Values}
Using only the reported numbers, the selected branched configuration changes the runtime from 0.6607 s to 0.6470 s while preserving the reported mAP of 0.680. The inference-time geometric strategy changes the mAP by +0.033 relative to the baseline, while its runtime changes by +0.0107 s. The trained geometric variant changes the mAP by +0.017 and the runtime by only +0.0001 s relative to the baseline. These are arithmetic differences computed from the thesis-reported values; they do not represent newly measured experiments.

\section{Analysis of the Proposed Design}
\subsection{Speed-Accuracy Decomposition}
The architecture is easier to interpret when speed and accuracy are considered as separate mechanisms. Branching reduces computation by allowing the network to stop early. Geometric features improve the evidence available for class decisions. Keeping the mechanisms separate also makes it possible to evaluate their effects independently.

The branch experiments show that an intermediate output is not automatically beneficial simply because it appears earlier. An exit must be placed at a depth where it can make a sufficiently useful prediction, otherwise the extra head can become overhead. This explains why the thesis performs several placement experiments and eventually keeps three outputs.

The cell-wise branching rule additionally demonstrates a distinction between \emph{image-level} and \emph{instance/cell-level} computation. In a multi-sign image, the desired computational path is not necessarily the same for every object. A simple sign can be accepted early while another sign continues deeper. The proposed cell-level formulation approximates this behavior within a single feature map.

\subsection{Why Geometric Features Help}
The geometric module supplies information that is structurally different from RGB appearance. Consider two signs that share a similar color distribution but have different shapes. A pixel-based representation can find both signs easy to localize while still confusing their classes. The template comparison adds an explicit test of shape consistency.

The thesis also demonstrates why the segmentation stage matters. A simple TV-plus-k-means pipeline was tested, but the thesis reports that, at approximately 450$\times$450 pixels, the simpler alternative was not effective enough for preserving the structure of small traffic signs. The Bayesian segmentation method was therefore retained for the geometric path.

\subsection{Inference-Time Versus Training-Time Fusion}
The two geometric-feature variants represent two deployment strategies. Inference-time verification leaves the main network untouched and applies a separate shape check to each detected box. This makes the method easy to retrofit to an existing model, but the extra comparison contributes to inference latency.

Training-time integration has the opposite profile. It increases the cost and complexity of training because the segmentation path must operate in parallel and its information must participate in optimization. In return, the final deployed network can use the learned geometric information without repeating the full auxiliary computation for every frame.

The thesis therefore suggests a general design principle: when the deployment environment has strict latency requirements, part of the computation may be moved from inference to training so long as the learned representation preserves the useful information.

\subsection{Data Quality and Class Balance}
The dataset-generation strategy is not merely a convenience. Since the detector must learn both localization and class identity, the class distribution and scene distribution matter simultaneously. Seamless cloning provides many combinations of scene and sign identity, while augmentation changes brightness, orientation, and affine geometry.

The thesis specifically notes that the distribution of these transformations should remain approximately uniform across classes. Otherwise, a model can become biased toward classes that are overrepresented under a particular condition. The data-generation pipeline thus functions as part of the recognition strategy rather than as an unrelated preprocessing step.

\subsection{Scope of the Reported Evaluation}
The conclusions reported here are limited to the experimental setup of the thesis: ten traffic-sign classes, the generated dataset, the stated training configuration, and the reported hardware/software environment. The numerical results should therefore be interpreted as experimental results for that setup. They do not establish performance across all geographic regions, camera systems, weather conditions, or complete traffic-sign taxonomies.

\section{Limitations and Future Directions}
Several limitations are identifiable from the thesis itself. First, the branched architecture increases training time because every branch must be trained. The thesis reports about ten days for the three-output model compared with about one week for the baseline, and longer times for configurations with more exits.

Second, the placement of intermediate exits is important. Outputs that are too close to one another can add computation without providing a proportionate benefit. A deeper branched architecture is proposed as future work, but it would require careful placement of exits and a network structure different from the standard YOLOv3 arrangement because the thesis argues that max-pooling should occur before the intermediate outputs.

Third, the geometric method can be further optimized for deployment. The thesis suggests shortening the geometric-feature extraction stage so that the method could also be combined with the branched detector. It also proposes combining the geometric-feature method with 2YOLOv-B and exploring a design in which intermediate outputs process only the relevant feature-map cells instead of every cell.

Fourth, the geometric thresholds are experimentally chosen. Extending the method to larger and more diverse traffic-sign collections would require re-evaluating the template representation and the empirical decision intervals.

A broader future direction is therefore a unified architecture in which early exits, cell selection, geometric verification, and learned feature fusion are jointly optimized. Such a model could potentially allocate computation adaptively across both space and network depth while retaining explicit shape information for fine-grained classification.

\section{Conclusion}
This paper presents a condensed research version of a traffic-sign recognition system developed as an M.Sc. thesis. The system uses YOLOv2 as a unified detector/classifier and introduces two complementary enhancements.

The first enhancement is 2YOLOv-B, a branched YOLOv2 architecture with intermediate outputs. The thesis evaluates several branch placements and retains a three-output structure. Whole-image branching provides the fastest reported runtime among the tested branching variants, while cell-wise branching addresses the problem of multiple signs with different recognition difficulty by allowing individual feature-map cells to terminate or continue independently.

The second enhancement is a geometric-feature mechanism based on unsupervised Bayesian image segmentation and class-specific binary templates. The segmented region inside the predicted bounding box is compared with the expected shape of the predicted class, including an inverted-template case to handle binary-label reversal. The approach can be used directly at inference time or incorporated during parallel training.

The reported experiments provide a compact view of the speed-accuracy trade-offs. YOLOv2 reports 0.6607 s runtime and 0.680 mAP. The selected three-output branched configuration reports 0.6470 s and 0.680 mAP. Geometric verification during inference reports 0.6714 s and 0.713 mAP, while the trained geometric-feature variant reports 0.6608 s and 0.697 mAP. The dataset component combines GTSDB and GTSRB through seamless cloning and controlled transformations to support simultaneous detection and classification.

Taken together, the thesis demonstrates a system-level strategy in which computational depth is adapted through branching and class evidence is strengthened through explicit geometry. The combination provides a compact framework for studying real-time traffic-sign recognition under a constrained inference budget.

\section*{Author and Thesis Information}
This manuscript is derived from the M.Sc. thesis titled \emph{Traffic Sign Recognition System based on Hierarchical Convolutional Neural Networks and Using Geometric Features For Autonomous Driving}, submitted to the Faculty of Computer Engineering, K. N. Toosi University of Technology, in Summer 2019. The thesis lists Dr.~Alireza Fatehi and Dr.~Behrooz Nasihatkon as supervisors.

\section*{Acknowledgments}
The author gratefully acknowledges the supervision and guidance of Dr.~Alireza Fatehi and Dr.~Behrooz Nasihatkon.

\paragraph{Language-assistance statement.}
An AI-based language tool was used to assist with English-language editing and restructuring of this manuscript. The author remains responsible for the scientific content, reported results, citations, and final version of the work.

\end{document}